%% file: main.tex
\documentclass{article}

\usepackage{spconf,graphicx}
\usepackage{amsmath,amssymb,mathrsfs,bbm}
\usepackage{graphicx}
\usepackage[colorlinks=true, allcolors=blue]{hyperref}
\usepackage{multirow}
\usepackage[lined,linesnumbered,ruled]{algorithm2e}

\usepackage{balance}

\usepackage{cite}

\usepackage{psfrag,epsfig,graphics}
\usepackage{amsmath,amsthm,amssymb,multirow}
\usepackage{mathbbol}
\usepackage{amssymb}   
\usepackage{threeparttable,balance}

\DeclareSymbolFontAlphabet{\amsmathbb}{AMSb}%

\input{MathSymbolDefs.tex}

\usepackage{color}  %

\title{Riemannian Change Point Detection on Manifolds \\with Robust Centroid Estimation}
\name{Xiuheng Wang~$^*$, Ricardo Borsoi~$^*$, Arnaud Breloy~$^\dag$, C\'edric Richard~$^\ddag$
\thanks{The work of Cédric Richard was supported in part by the French Government through the 3IA Côte d'Azur Investments in the Future Project under grant ANR-19-P3IA-0002, and in part by grant ANR-19-CE48-0002. 
The work of Ricardo Borsoi was supported in part by the French National Research Agency, under grants ANR-23-CE23-0024, ANR-23-CE94-0001, and by the National Science Foundation, under grant NSF 2316420.}
}
\address{\small $^*$ Universit\'e de Lorraine, CNRS, CRAN, France \\
        \small
        $^\dag$ Conservatoire National des Arts et M\'etiers, CEDRIC, France\\
        \small  $^\ddag$ Universit\'e C\^ote d'Azur, CNRS, OCA, France\\
	\small dr.xiuheng.wang@gmail.com, ricardo.borsoi@univ-lorraine.fr, 
    arnaud.breloy@lecnam.net, cedric.richard@unice.fr}

\begin{document}
\maketitle
\begin{abstract}
Non-parametric change-point detection in streaming time series data is a long-standing challenge in signal processing. Recent advancements in statistics and machine learning have increasingly addressed this problem for data residing on Riemannian manifolds. One prominent strategy involves monitoring abrupt changes in the center of mass of the time series. Implemented in a streaming fashion, this strategy, however, requires careful step size tuning when computing the updates of the center of mass. In this paper, we propose to leverage robust centroid on manifolds from M-estimation theory to address this issue. Our proposal consists of comparing two centroid estimates: the classical Karcher mean (sensitive to change) versus one defined from Huber's function (robust to change). This comparison leads to the definition of a test statistic whose performance is less sensitive to the underlying estimation method. We propose a stochastic Riemannian optimization algorithm to estimate both robust centroids efficiently. Experiments conducted on both simulated and real-world data across two representative manifolds demonstrate the superior performance of our proposed method.
\end{abstract}
\begin{keywords}
Non-parametric, change point detection, Riemannian manifolds, robust centroid estimation, M-estimator, online, stochastic optimization.
\end{keywords}

\section{Introduction}
\label{sec:intro}

\paragraph*{Parametric and non-parametric CPD:} 
The change-point detection (CPD) problem aims to identify abrupt changes in the statistical distribution of time series data. CPD has an extensive history within the statistics literature, with diverse applications ranging from medical~\cite{gajic2015detection} and speech processing~\cite{rybach2009audio} to image analysis~\cite{borsoi2021online}. Various CPD algorithms have been proposed depending on the level of prior information available. Classical CPD approaches typically utilize parametric strategies, requiring knowledge of the specific form of the probability density function (PDF) before and after the change point. Notable parametric CPD algorithms include the cumulative sum (CUSUM)~\cite{page1954continuousInspectionCUSUM}, designed for detecting shifts in either the mean or variance, and the generalized likelihood ratio test (GLRT)~\cite{gustafsson1996marginalized}, which models the data PDF using a linear state-space representation.

While strategies based on parametric statistical models can offer strong performance guarantees, non-parametric settings have been the subject of increasing interest since they do not make strict assumptions on the data distribution that could make the algorithm sensitive to modeling errors~\cite{truong2020selectiveReviewOfflineCPD}.
Notable approaches include the Exponentially Weighted Moving Average (EWMA)~\cite{costa2006single} and kernel Maximum Mean Discrepancy (MMD)~\cite{gretton2006kernel}. Recently, the NEWMA algorithm~\cite{keriven2020newma} was introduced. It detects change points by comparing two EWMAs streaming statistics~\cite{costa2006single}, each computed with different forgetting factors. 
The kernel MMD statistic, originally introduced for hypothesis testing in~\cite{gretton2006kernel}, has also found extensive application in kernel CPD~\cite{harchaoui2008kernel}.
Another kernel-based algorithm, which makes use of adaptive density ratio estimation, was developed in~\cite{ferrari2022online}.  
Finally, the potential of neural networks was investigated in~\cite{wang2023_NODE}.

\paragraph*{CPD on manifolds:}
The extension of nonparametric CPD algorithms to handle data defined on non-Euclidean spaces, such as Riemannian manifolds, has been the subject of recent research~\cite{dubey2020frechet,wang2023online, wang2024nonparametric}. Formally, given a time series of independent random variables $\{\bx_t\}_{t\in\bbN}$ lying on a Riemannian manifold $\calM$, the Riemannian CPD problem~\cite{wang2023online,wang2024nonparametric} consists of detecting a time index $t_r\in\bbN$ at which the distribution of the data abruptly changes:
\begin{equation}
    \label{eq:initial_problem}
	t<t_r:\; \bx_{t} \sim P_1(\bx)\,, \qquad\;t\geq t_r:\ \bx_{t} \sim P_2(\bx),
\end{equation}
where $P_1(\bx)$ and $P_2(\bx)$ denote the probability measure of the data $\bx_t\in\calM$ before and after the change point $t_r$, respectively. A generalization of problem~\eqref{eq:initial_problem} is the \emph{multiple CPD problem}, where one aims to estimate not a single but a set of change points $\{t_{r_1},t_{r_2},\dots\}\subset\amsmathbb{N}_*$, meaning that $\bx_{t}$ is allowed to change at multiple locations.

\paragraph*{Existing algorithms and contribution:}
Because they do not account for the geometry of the data residing on $\calM$, particularly the absence of a vector space structure, standard nonparametric algorithms are not suitable for addressing the problem~\eqref{eq:initial_problem}.
For manifold-valued data, recent works have proposed to monitor abrupt changes in the Karcher mean~\cite{karcher1977riemannian} on $\calM$, defined as:
\begin{equation}
    \label{eq:optimization_karcher}
	\bm^* = \mathop{\arg\min}_{\bm\in\calM}\, f(\bm) \,,
\end{equation}
which is the point that minimizes the Karcher variance: \[f(\bm)  = \bbE_{\bx\sim P(\bx)} \left\{\frac{1}{2}d_{\calM}^2(\bm, \bx)\right\}.\]
For example, an offline technique monitoring changes in $\bm^*$ before and after change points was proposed in~\cite{dubey2020frechet}. 
More recently, an online algorithm~\cite{wang2023online, wang2024nonparametric} extended NEWMA to general manifolds by comparing two recursive estimates of $\bm^*$ computed via stochastic gradient algorithms. However, none of these methods considered a more general alternative: 
recursively monitoring two \emph{different} statistics of the data $\bx_t$ constructed to be similar in the absence of a change point and significantly different after a change point occurs. 

A key concept in our work is to leverage a robust centroid estimator on $\calM$. Based on the theory of M-estimators, a class of robust centroid estimators arises from the so-called Huber function~\cite{huber1964robust, tyler1987distribution} and covers the concept of Karcher mean. Robust centroid estimation has recently been applied to various tasks in manifolds~\cite{ilea2016m, breloy2021majorization}.
The primary contribution of this paper is the introduction of robust centroid
estimation into Riemannian CPD. We specially explore two special cases of the Huber centroid, one prioritizing adaptability and the other emphasizing robustness. The comparison between these two centroids forms the basis for our test statistic, which notably reduces the dependence on the convergence of the estimation algorithm. Additionally, we propose an online estimation method for the Huber centroid using a Riemannian stochastic optimization algorithm. Our methodology is tailored explicitly for two commonly encountered manifolds: the symmetric positive definite (SPD) manifold and the Grassmann manifold. Numerical experiments on both synthetic and real-world datasets validate the effectiveness of our Riemannian CPD approach.

\section{Background}
\label{sec:background}
This section briefly introduces concepts of Riemannian geometry and optimization on manifolds~\cite{absil2009manoptBook, boumal2023introduction}.

A \textit{Riemannian manifold} $(\calM, g)$ is a constrained set $\calM$ endowed with a \textit{Riemannian metric} $g_x(\cdot, \cdot):T_x\calM\times T_x\calM\to\bbR$, defined for every point $x\in\calM$, with $T_x\calM$ the so-called \textit{tangent space} of $\calM$ at $x$. 
A \textit{geodesic} $\gamma_v:[0,1]\to\calM$ is the minimal length curve linking two points $x,y\in\calM$ such that $x=\gamma_v(0)$ and $y=\gamma_v(1)$, with $v\in T_x\calM$ the velocity of $\gamma_v$ at $0$ denoted by $\dot{\gamma}_v(0)$. The geodesic distance \textit{} $d_{\calM}(\cdot\,,\cdot):\calM\times\calM\to\bbR$ is defined as the length of the geodesic that connects two points $x,y\in\calM$. 
The \textit{exponential map} $w=\exp_{x}(v)$ is defined as the point $w\in\calM$ located on the unique geodesic $\gamma_v(t)$ with endpoints $x=\gamma_v(0)$, $w=\gamma_v(1)$ and velocity $v=\dot{\gamma}_v(0)$. 
Calculating the exponential map can be computationally demanding. 
In practice, it is common to employ a \textit{retraction} $R_{x}: T_{x}\calM\to\calM$ instead, defined at every ${x}\in\calM$, which consists of a second-order approximation to the exponential map, satisfying $d_{\calM}(R_{x}(tv),\exp_{{x}}(tv))=O(t^3)$.
Consider a smooth function $f:\calM \to \bbR$. The \textit{Riemannian gradient} of $f$ at $x\in\calM$ is defined as the unique tangent vector $\nabla f(x)\in T_x\calM$ satisfying
	$\frac{d}{dt}\big|_{t=0}f(\exp_{x}(tv)) = \langle\nabla f(x), v\rangle_x$,
for all $v\in T_x\calM$.

\section{Methodology}
\label{sec:method}

In the online setting, samples $\bx_t$ in the Riemannian manifold $\calM$ are observed sequentially over time. Thus, at each time instant $t'\in\amsmathbb{N}_*$ it is necessary to determine whether to flag $t'$ as a changepoint based on previously measured data, $\{\bx_1,\dots,\bx_{t'}\}$. Considering the signal model~\eqref{eq:initial_problem} in which a change point occurs at time $t_r$, the detection problem in our setting involves two distinct objectives, the first being to minimize the mean detection delay (MDD) $\amsmathbb{E}\{\hat{t}_r-t_r\}$, with $\hat{t}_r$ being the first detection after the true change point $t_r$, while the second is to maximize the average run time (ARL) $\amsmathbb{E}\{\hat{t}_r\}$, where $\hat{t}_r<t_t$ denotes the time at which the first false alarm is flagged.

\subsection{Robust centroid estimation}
To detect change points on $\calM$, we propose monitoring the robust centroid of the data stream $\bx_t\in\calM$, extending the framework of the Karcher mean defined in~\eqref{eq:optimization_karcher}.
Based on the theory of M-estimation, the robust centroid can be obtained by minimizing the following cost function:
\begin{equation}
    \label{eq:optimization}
	\bm_{\rho}^* = \mathop{\arg\min}_{\bm_{\rho}\in\calM}\, f_{\rho}(\bm_{\rho}) \,,
\end{equation}
where \[f_{\rho}(\bm_{\rho})  = \bbE_{\bx\sim P(\bx)} \left\{\frac{1}{2}\rho(d_{\calM}(\bm_{\rho}, \bx))d_{\calM}^2(\bm_{\rho}, \bx)\right\},\]
with $\rho:\amsmathbb{R}\to\amsmathbb{R}$ a function
that ensures the robustness to outliers.

Depending on $\rho$, various robust centroid estimators can be defined~\cite{breloy2021majorization}. 
For example, the Huber's estimator~\cite{huber1964robust} can be considered with a specific Huber function defined as
\begin{equation}
    \label{eq:huber}
    \rho(a) = \min\left(1, \frac{A}{a}\right)\,,
\end{equation}
where $A > 0$ is a fixed parameter. 

In~\cite{wang2023online, wang2024nonparametric}, the authors consider only the Karcher mean estimator defined in~\eqref{eq:optimization_karcher}. In this work, we extend this strategy by viewing the Karcher mean estimator as a special instance of the robust centroid estimator in~\eqref{eq:optimization} corresponding to the choice of $A=\infty$ in~\eqref{eq:huber}. We further compare this adaptive centroid estimator with another special case of the robust centroid estimator obtained by setting $A\in(0, \infty)$ in~\eqref{eq:huber}, which provides enhanced robustness in the presence of outliers -- in our context, the post-change samples. 

Although other robust centroid estimators employing different forms of $\rho$ exist, such as $\ell_1$-norm, Cauchy–Lorentz-type and Geman-McClure-type~\cite{breloy2021majorization}, we restrict our analysis to Huber-type estimators. This choice preserves geodesic convexity and smoothness of the cost function on specific manifolds, notably the SPD manifold. This allows us to leverage the powerful Riemannian stochastic gradient descent (SGD)~\cite{bonnabel2013stochasticGradRiemannian, zhang2016firstOrderGeodesicallyConvex} algorithm, whose convergence properties strongly rely on these characteristics of the cost function.

\subsection{Riemannian SGD-based estimates}
As discussed above, the proposed Riemannian CPD strategy monitors abrupt changes in the robust centroid of the data stream. A key requirement is that change points must be detected \emph{online}, relying solely on past data. In \cite{wang2023online, wang2024nonparametric}, within the Riemannian geometry framework, Karcher mean estimates were computed using two Riemannian SGD with distinct step sizes. However, this approach can be restrictive, as these step sizes must be selected within a limited range to ensure the convergence of the Riemannian SGD (see Theorem~4.1 in~\cite{wang2024nonparametric}). To address this issue, we propose comparing two estimates derived from two distinct special cases of the robust centroid estimator, both computed using the \emph{same} step size, as follows. 

On the one hand, when choosing $A = \infty$ in~\eqref{eq:huber}, the robust centroid estimator in~\eqref{eq:optimization} reduces to the Karcher mean defined in~\eqref{eq:optimization_karcher}. Therefore, for this case, we consider the following iterative update using Riemannian SGD~\cite{bonnabel2013stochasticGradRiemannian, zhang2016firstOrderGeodesicallyConvex}:
\begin{align}
\label{eq:optimization_alpha}
	\bm_{t+1} &= R_{\bm_{t}}\big(-\alpha H(\bm_{t}, \bx_t)\big) \,,
\end{align}
where $\alpha$ is the step size and $H(\bm_{t}, \bx_t)$ is the stochastic Riemannian gradient of $f(\bm_{t})$, approximated using the streaming data $\bx_t$. This yields a baseline estimator that, according to the theory of M-estimation, is less robust to outliers and, consequently, to change points in an online setting.

On the other hand, when $A\in(0, \infty)$, the robust centroid estimator defined in~\eqref{eq:optimization} can be computed using Riemannian SGD through the following iterative update:
\begin{align}
    \label{eq:optimization_alpha_rho}
	\bm_{\rho, t+1} &= R_{\bm_{\rho,t}}\big(-\alpha H_\rho(\bm_{\rho, t}, \bx_t)\big) \,,
\end{align}
where $H_\rho(\bm_{\rho, t}, \bx_t)$ is the stochastic Riemannian gradient of $f_{\rho}(\bm_{t})$ approximated using the streaming data $\bx_t$.
Using the Huber function defined in~\eqref{eq:huber}, the relationship between $H_\rho(\bm_{\rho, t}, \bx_t)$ and $H(\bm_{\rho, t}, \bx_t)$ can be derived as follows:
\begin{align}
    \label{eq:relation}
    &H_{\rho}(\bm_{\rho, t}, \bx_t)  = \nonumber \\
    &\quad\,\,\,\begin{cases} H(\bm_{\rho, t}, \bx_t), & d_{\calM}(\bm_{\rho, t}, \bx_t) \leq A, \\  \frac{A}{d_{\calM}(\bm_{\rho, t}, \bx_t)}H(\bm_{\rho, t}, \bx_t) , & d_{\calM}(\bm_{\rho, t}, \bx_t) > A. \end{cases}
\end{align}
This result indicates that the stochastic gradient matches that of the Karcher mean estimator when data points are close to the current centroid estimator (i.e., $d_{\calM}(\bm_{\rho, t}, \bx_t) \leq A$), but it is scaled down for distant ones (i.e., $d_{\calM}(\bm_{\rho, t}, \bx_t) > A$). Thus, the gradient becomes more robust against outliers and, consequently, less sensitive to abrupt changes in the online setting.

\subsection{An adaptive CPD statistic} 
\label{ssec:adaptive}

\begin{algorithm} [t]
\small
\SetKwInOut{Input}{Input}
\caption{Riemannian CPD with robust centroid estimates~\label{alg:global_alg}}
\Input{\mbox{$\{\bx_t\}$, step sizes $\alpha$, threshold~$\xi$.}}
Initialization: $\bm_{0} = \bm_{\rho, 0} = \bx_0$ \;
\For{$t=1,2,3,\ldots$}{
Update the adaptive and robust centroid estimates $\bm_{t}$ and $\bm_{\rho, t}$ using~\eqref{eq:optimization_alpha} and~\eqref{eq:optimization_alpha_rho}\;
Compute the test statistic $g_t = d_{\calM}(\bm_{t}, \bm_{\rho,t})$ \;
\If{$ g_t > \xi$}{Flag $t$ as a change point\;}
}
\end{algorithm}

Using these two estimates $\bm_{t}$ and $\bm_{\rho, t}$, we can define an adaptive CPD statistic by comparing them with geodesic distance on $\calM$ as:
\begin{equation}
    \label{eq:cpd_statistic}
	g_t = d_{\calM}(\bm_{t}, \bm_{\rho,t}) \,.
\end{equation}
CPD is then performed by comparing $g_t$ to a threshold $\xi$. The full CPD procedure is summarized in Algorithm~\ref{alg:global_alg}.

The statistic $g_t$ can be interpreted as a metric for detecting abrupt changes in streaming data on the manifold $\calM$. Specifically, $\bm_{t}$ and $\bm_{\rho,t}$ serve as adaptive estimators tracking the standard and robust centroids, respectively. %

When a change point occurs within sequence $\{1, \dots, T\}$, the underlying data distribution shifts, causing measurements $\{x_1, \dots, x_T\}$ to include data points drawn from a different distribution $P_2(\bx)$ which can be viewed as outliers. The non-robust estimator $\bm_{t+1}$ adapts gradually to this new regime. In contrast, the robust estimator $\bm_{\rho, t+1}$ remains stable against transient shifts and exhibits significant adaptation only when the contamination surpasses its robustness threshold. 
This behavior highlights the benefit of robust centroid estimation for Riemannian CPD, as lower sensitivity to abrupt changes makes the robust estimator an ideal baseline in constructing effective test statistic.

\section{Applications and experiments}
\label{sec:experiment}
In this section, we apply Algorithm~\ref{alg:global_alg} to tackle two common instances of manifolds: the SPD manifold of $p\times p$ matrices, denoted by $\SPD$, and the Grassmann manifold, representing the set of $k$-dimensional linear subspaces of $\bbR^p$, denoted by $\GR$.
The implementation of Algorithm~\ref{alg:global_alg} on these manifolds requires manifold-specific definitions of the geodesic distance $d_{\calM}(\bm, \bx)$, the stochastic Riemannian gradients $H(\bm, \bx)$ and $H_{\rho}(\bm, \bx)$, and a second-order retraction operator $R_{\bm}$. The definitions of $d_{\calM}(\bm, \bx)$, $H(\bm, \bx)$ and $R_{\bm}$ for $\SPD$ and $\GR$ are provided in Section $5$ of~\cite{wang2024nonparametric}. The corresponding robust gradient $H_{\rho}(\bm, \bx)$ can be derived using the relation defined in~\eqref{eq:relation} based on $H(\bm, \bx)$.

\begin{figure}[t]
    \centering
    \includegraphics[trim = 2mm 2mm 2mm 2mm, clip, scale=0.54]{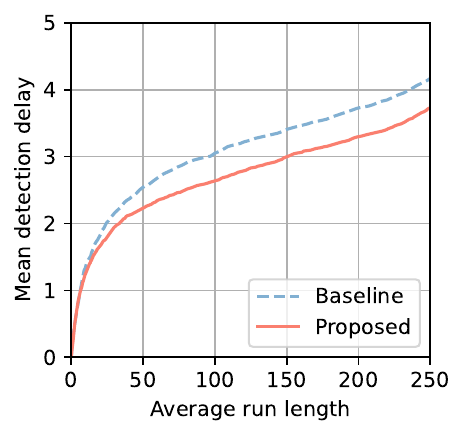} 
    \includegraphics[trim = 2mm 2mm 2mm 2mm, clip, scale=0.54]{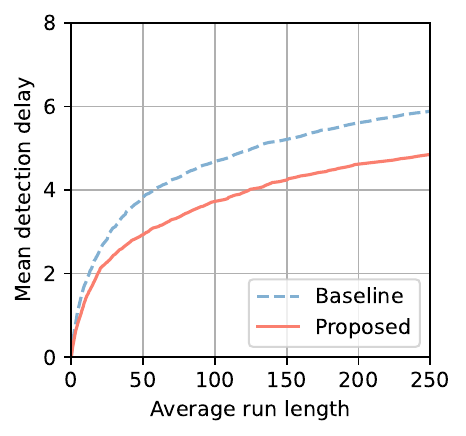} 
    \caption{ARL versus MDD for the compared algorithms on synthetic data on both $\SPD$ (left) and $\GR$ (right)..}
    \label{fig:synthetic}
\end{figure}

\begin{figure}[t]
    \centering
    \includegraphics[trim = 2mm 2mm 2mm 2mm, clip, scale=0.54]{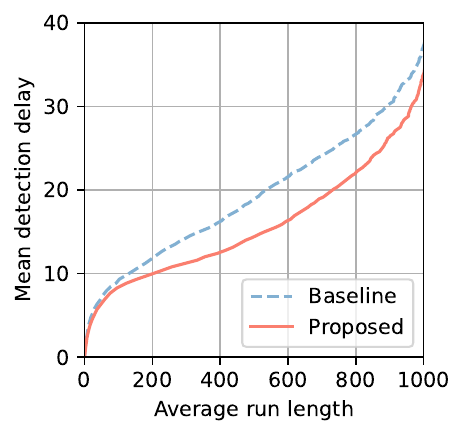} 
    \includegraphics[trim = 2mm 2mm 2mm 2mm, clip, scale=0.54]{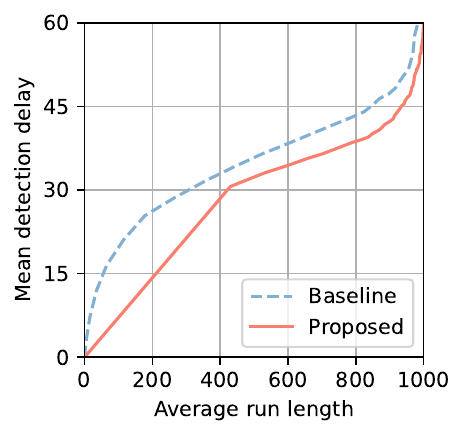} 
    \caption{ARL versus MDD for the compared algorithms on real data on both $\SPD$ (left) and $\GR$ (right).}
    \label{fig:real}
\end{figure}

We shall now illustrate the performance of Algorithm~\ref{alg:global_alg} for Riemannian CPD on both manifolds $\SPD$ and $\GR$. 
For Algorithm~\ref{alg:global_alg}, the step size was set to $\alpha = 0.05$, and the parameter $A$ was set to $1$ for $\SPD$ and $0.05$ for $\GR$.  
We selected a non-parametric online algorithm for Riemannian CPD~\cite{wang2023online, wang2024nonparametric} as a baseline for comparison with Algorithm~\ref{alg:global_alg}.
To compare the detection performance of these algorithms, $10^4$ Monte Carlo simulations were performed to estimate the MDD and ARL for both synthetic and real data. Open-source code to reproduce the results is available on \url{https://github.com/xiuheng-wang/CPD_RCE_release}.

\subsection{Experiment with synthetic data:} 
We first present results obtained from synthetically generated sequences of i.i.d. data on $\SPD$ and $\GR$. \vspace{-0.3cm}

\paragraph*{SPD Manifold:}  
We sampled matrices $\bSigma_t \in \SPD$ with $p = 10$ from a Wishart distribution with scaling matrix $\bV$ and $p+2$ degrees of freedom. A total of 2000 samples were generated, with a change point at $t_r = 1500$, where $\bV$ was reset.  \vspace{-0.3cm}

\paragraph*{Grassmann Manifold:}  
The data $\pi(\bU_{\!t}) \in \GR$ with $p = 20$ and $k = 5$ was generated in two steps. First, we sampled matrices $\bZ_t$ from a random matrix Gaussian distribution. Then, the orthonormal matrices $\bU_{\!t}$ were obtained as the left singular vectors corresponding to the $k$ largest singular values of $\bZ_t$. A total of 2000 samples was generated, with a change point at $t_r = 1500$, where the mean of the matrix Gaussian distribution of $\bZ_t$ was reset. \vspace{-0.3cm}

\paragraph*{Results:}
The MDD as a function of ARL for all methods is depicted in Fig.~\ref{fig:synthetic} for both manifolds. It is evident that Algorithm~\ref{alg:global_alg} results in a lower MDD for a fixed ARL when compared to the baseline.

\subsection{Experiment with real data:} 
We now present results obtained from real data on both $\SPD$ and $\GR$, focusing on the task of detecting speech segments within noisy backgrounds. 
We first combined 4 seconds of real speech from the TIMIT database~\cite{garofolo1993timit} with 15 seconds of background noise recorded in real street environments from the QUT-NOISE database~\cite{dean2010qut}, ensuring a Signal-to-Noise Ratio (SNR) of -3 dB. Next, we applied the Short-Time Fourier Transform (STFT)~\cite{cohen1995time} to the resulting one-dimensional audio signal, thereby obtaining frequency-domain representations as a $d = 128$ dimensional time series, $\bs_t \in \bbR^d$. Finally, we averaged neighboring frequency channels of $\bs_t$ to produce a downsampled representation with reduced dimensionality of 16 channels. \vspace{-0.3cm}

\paragraph*{SPD Manifold:} 
We generated data points $\bSigma_t \in \SPD$ with $p = 16$ by computing covariance matrices from sliding windows, each containing 32 consecutive samples. \vspace{-0.3cm}

\paragraph*{Grassmann Manifold:} 
We applied truncated singular value decomposition (SVD), retaining only the first singular vector (\(k = 1\)), to the samples within each sliding window. This yielded orthonormal matrices $\bU_{\!t}$ that define the subspaces $\pi(\bU_{\!t}) \in \GR$. \vspace{-0.3cm}

\paragraph*{Results:}
The performance curves for all methods are represented in Fig.~\ref{fig:real}. It is important to emphasize that this problem setting is particularly challenging due to the complexity of real acoustic signals and the non-i.i.d. nature of the extracted features.
Nevertheless, Algorithm~\ref{alg:global_alg} consistently demonstrates superior performance in terms of  MDD versus ARL compared to the baseline methods.

\section{Conclusion}
\label{sec:conclusion}
In this paper, we introduced a robust centroid estimation approach for Riemannian CPD. We proposed an adaptive test statistic comparing two specialized robust centroid estimators computed via Riemannian SGD: one emphasizing adaptability to new data, and the other enhancing robustness against abrupt changes. Importantly, our CPD statistic requires selecting only a single step size, a crucial advantage given that Riemannian SGD convergence is guaranteed only within a limited step-size range. Experimental results on both SPD and Grassmann manifolds confirm that our method consistently outperforms the previous approach based on two Karcher mean estimators.

\ninept
\bibliographystyle{IEEEbib}
\balance
\bibliography{manifoldOptRefs, NodeRefs, RubostEst}

\end{document}

%% file: MathSymbolDefs.tex
\newcommand{\bU}{\boldsymbol{U}}
\newcommand{\bV}{\boldsymbol{V}}

\newcommand{\bZ}{\boldsymbol{Z}}

\newcommand{\bm}{\boldsymbol{m}}

\newcommand{\bs}{\boldsymbol{s}}

\newcommand{\bx}{\boldsymbol{x}}

\newcommand{\calG}{\mathcal{G}}

\newcommand{\calM}{\mathcal{M}}

\newcommand{\bbE}{\amsmathbb{E}}

\newcommand{\bbN}{\amsmathbb{N}}

\newcommand{\bbR}{\amsmathbb{R}}

\newcommand{\SPD}{\mathcal{S}_p^{++}}
\newcommand{\GR}{\calG_p^k}

\newcommand{\bSigma}{\boldsymbol{\Sigma}}

\usepackage{scalerel,stackengine}